\documentclass[10pt,twocolumn,letterpaper]{article}

\usepackage{cvpr}              

\usepackage{times}
\usepackage{epsfig}
\usepackage{graphicx}
\usepackage{amsmath}
\usepackage{amssymb}

\definecolor{cvprblue}{rgb}{0.21,0.49,0.74}
\usepackage[pagebackref,breaklinks,colorlinks,allcolors=cvprblue]{hyperref}

\def\paperID{17} 
\def\confName{CVPR}
\def\confYear{2025}

\definecolor{grayhighlight}{RGB}{213,229,255}

\newsavebox\CBox

\begin{document}
	
	\title{RGB Photo Enhancement with Deep Learning on Smartphone GPUs: \\ Mobile AI 2026 Challenge Report}
	
	\author{
		Andrey Ignatov \and Georgy Perevozchikov \and Radu Timofte 
		\and Dong Zhang \and Zhao Ran \and Yang Zhang \and Xiaochen Li \and Shichang Ju \and Yiqin Cao \and Yang Lu
		\and Nguyen Nhu Tinh Anh \and To Thanh Dat \and Nguyen Ngoc Ky Duyen \and Nguyen Trong Nghia \and Bui Minh Hieu
		\and Qiuhai Yan \and Sunzhe Yang \and Kang Chen \and Tingting Wang \and Faming Fang
		\and Runhua Deng \and Zhifeng Wang \and Long Peng \and Aiwen Jiang
	}
	
	\maketitle

	\begin{abstract}
		 
		  On-device RGB photo enhancement is a popular task that is needed for quality improvement of photos captured with mobile devices as well as for retouching of images acquired through other sources.
		  This creates a need for efficient deep learning models for this problem that are not only performant but can also be deployed on constrained mobile hardware. In this Mobile AI challenge, we address this problem and propose participants to design fast solutions that can perform comprehensive enhancement of RGB photos while running on popular smartphone GPUs from Qualcomm and Arm. For this, the participants were provided with a large-scale DPED dataset consisting of RGB image pairs captured with an old iPhone 3GS smartphone and a professional Canon 70D DSLR camera. The runtime of all models was evaluated on the Snapdragon 8 Elite Gen 5 and Dimensity 9500 chipsets with the Adreno and Mali GPUs, respectively. Top solutions proposed in this challenge are able to enhance Full HD resolution RGB images on both platforms under 300 milliseconds. A comprehensive description of the models developed in the challenge is provided in this paper.
		
	\end{abstract}
	{\let\thefootnote\relax\footnotetext{%
			\hspace{-5mm}$^*$Andrey Ignatov, Georgy Perevozchikov and Radu Timofte are the organizers of the Mobile AI 2026 challenge \textit{(andrey@vision.ee.ethz.ch, georgii.perevozchikov@uni-wuerzburg.de, radu.timofte@uni-wuerzburg. de)}. The other authors participated in the challenge. \\ Appendix \ref{sec:apd:team} contains the authors' team names and affiliations.\\ \\ Mobile AI 2026 Workshop website: \\ \url{https://ai-benchmark.com/workshops/mai/2026/}
	}}
	
	\section{Introduction}
	
\begin{figure*}[t!]
	\centering
	\resizebox{0.9\linewidth}{!}
	{
		\includegraphics[width=1.0\linewidth]{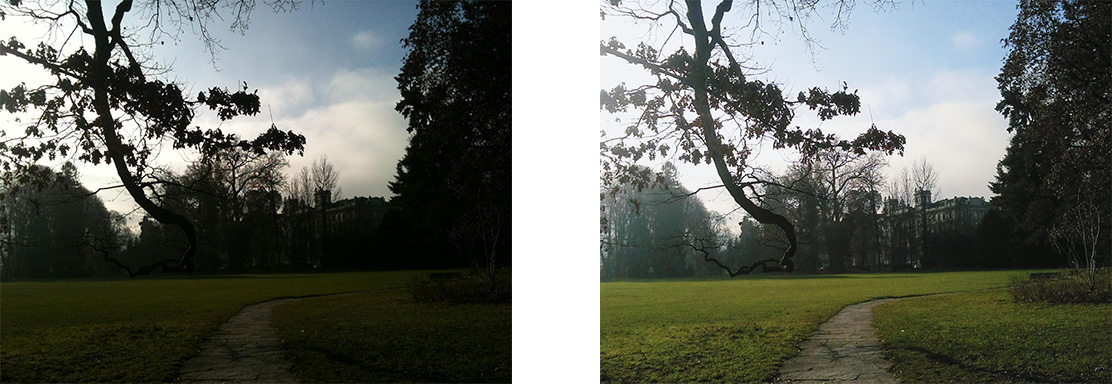}
	}
	\caption{\small{The original iPhone 3GS photo (left) and the same image enhanced by the DPED network~\cite{ignatov2017dslr} (right).}}
	\label{fig:example_photos}
\end{figure*}
	
	Image restoration and enhancement are fundamental problems in computer vision that aim to improve various aspects of image quality, including denoising, deblurring, super-resolution, color correction, contrast enhancement, dynamic range reconstruction and artifact removal. These research areas have attracted significant attention from the computer vision and graphics communities in recent years due to their broad practical relevance and inherent technical challenges. With the rapid growth of mobile photography, enhancing images captured by compact mobile camera sensors has become an active area of research, driving the development of advanced image enhancement techniques for smartphones and other resource-constrained devices. Tackling this problem as an end-to-end photo enhancement task was first proposed in~\cite{ignatov2017dslr,ignatov2018wespe}, where the authors presented a deep learning model for mapping low-quality smartphone photos to superior-quality images obtained with a high-end DSLR camera. The collected DPED dataset was later used in a number of  subsequent challenges~\cite{ignatov2018pirm,ignatov2019ntire} and works~\cite{vu2018fast,lugmayr2019unsupervised,de2018fast,hui2018perception,huang2018range,liu2018deep} that proposed numerous performant solutions for this problem. In this competition, we use the same dataset and propose the participants to design even more efficient RGB image enhancement models that can run on recent smartphone GPUs.
		
	Designing deep learning models that run efficiently on mobile devices~--- especially across different AI accelerators like NPUs~--- is a complex challenge. It demands careful analysis of hardware constraints such as limited operator support, restricted memory capacity and bandwidth, reduced accuracy for quantized models and specific driver behavior of NPUs. Even when certain layers or operations are technically supported, they may be executed inefficiently due to constraints related to NPU hardware architecture or its driver implementation, and thus create performance bottlenecks during inference. A striking example here would be the Google Tensor NPU present in all recent Google Pixel smartphones that, despite having quite high theoretical peak performance, is unable to efficiently accelerate the vast majority of even simplistic CNN-based models~\cite{ignatov2025MAISR}. As a result, achieving low latency and energy efficiency often requires partial or complete redesigns of model architectures. For a more detailed discussion of these challenges and the performance of modern mobile AI accelerators we refer to~\cite{ignatov2019ai,ignatov2018ai}. Overall, all these limitations require various adaptations of the standard deep learning models for their efficient deployment on mobile NPUs, including conventional optimizations like 16-bit / 8-bit~\cite{chiang2020deploying,jain2019trained,jacob2018quantization,yang2019quantization,li2021mqbench,nagel2021white,bondarenko2021understanding,siddegowda2022neural,tan2024mobilequant,lin2025awq} and low-bit~\cite{cai2020zeroq,uhlich2019mixed,banner2019post,ignatov2020controlling,liu2018bi,zhuo2022empirical,bhalgat2020lsq+,kim2023memory,dettmers2023case,bondarenko2024low,li2025memory} quantization, network pruning and compression~\cite{chiang2020deploying,li2019learning,
		liu2019metapruning,obukhov2020t,kim2021pqk,hawks2021ps,zheng2022model,xu2023sparsebytenn,kuzmin2023pruning,wang2024towards}, NPU-specific adaptations, platform-aware neural architecture search~\cite{howard2019searching,tan2019mnasnet,wu2019fbnet,
		wan2020fbnetv2,mills2021profiling,chu2021discovering,zhan2021achieving,wu2022compiler,chitty2022neural,yoshihama2023hardware}, \etc.
	
	In this Mobile AI Challenge, the participants were proposed to develop efficient deep learning models that can enhance Full HD resolution RGB images while running on smartphone GPUs. For this, the participating teams were provided with the DPED dataset~\cite{ignatov2017dslr} consisting of thousands of paired photos captured with a low-end 3.15MP iPhone camera and a professional Canon 70D DSLR. The runtime of the models was evaluated on mobile GPUs present in the Snapdragon 8 Elite Gen 5 and the Dimensity 9500 chipsets. Each submission was evaluated based on both image fidelity and runtime efficiency, thus balancing reconstruction quality with model efficiency. All solutions developed during this challenge are fully compatible with the LiteRT~\cite{LiteRT2025} (former TensorFlow Lite) framework, enabling deployment on a variety of Linux and Android-based IoT platforms, smartphones, and edge devices.
	
	\smallskip
	
	
	This challenge is a part of the \textit{MAI 2026 Workshop and Challenges} consisting of the following competitions:
	
	
	
	\begin{itemize}
		\item 4K Image Super-Resolution on Mobile NPUs~\cite{ignatov2025sr4k}
		\item Real-Time Video Super-Resolution on Smartphone GPUs~\cite{ignatov2025srvideo}
		\item Efficient Image Denoising on Smartphone GPUs~\cite{ignatov20254denoising}
		\item RGB Photo Enhancement on Smartphone GPUs
		\item Learned Smartphone ISP on Mobile GPUs~\cite{ignatov2025isp}
		\item Bokeh Effect Rendering on Smartphone GPUs~\cite{ignatov2025bokeh}
		\item Efficient LLMs for On-Device Deployment
		\item Adapting Stable Diffusion for Efficient On-Device Inference
	\end{itemize}
	
	\normalsize
	
	
	\noindent The results and solutions obtained in the previous \textit{Mobile AI Challenges} are described in the following papers:
	
	
	\begin{itemize}
		\item Image Super-Resolution on Mobile NPUs~\cite{ignatov2021real,ignatov2022efficient,ignatov2025MAISR}
		\item Learned ISP on Mobile NPUs and GPUs~\cite{ignatov2021learned,ignatov2022learned,ignatov2025MAIRAW}
		\item Video Super-Resolution on Mobile GPUs~\cite{romero2021real,ignatov2022power}
		\item Image Denoising on Mobile GPUs~\cite{ignatov2021fastDenoising}
		\item Bokeh Effect Rendering on Mobile GPUs~\cite{ignatov2022realistic}
		\item Camera Scene Detection on Smartphones~\cite{ignatov2021fastSceneDetection,pouget2021fast}
		\item RGB Photo Enhancement on Mobile GPUs~\cite{ignatov2025MAIRGB}
		\item Depth Estimation on Mobile Devices~\cite{ignatov2021fastDepth,ignatov2022efficient}
	\end{itemize}
	
	\normalsize
	
	\begin{figure*}[t!]
		\centering
		\setlength{\tabcolsep}{1pt}
		\resizebox{0.96\linewidth}{!}
		{
			\includegraphics[width=1.0\linewidth]{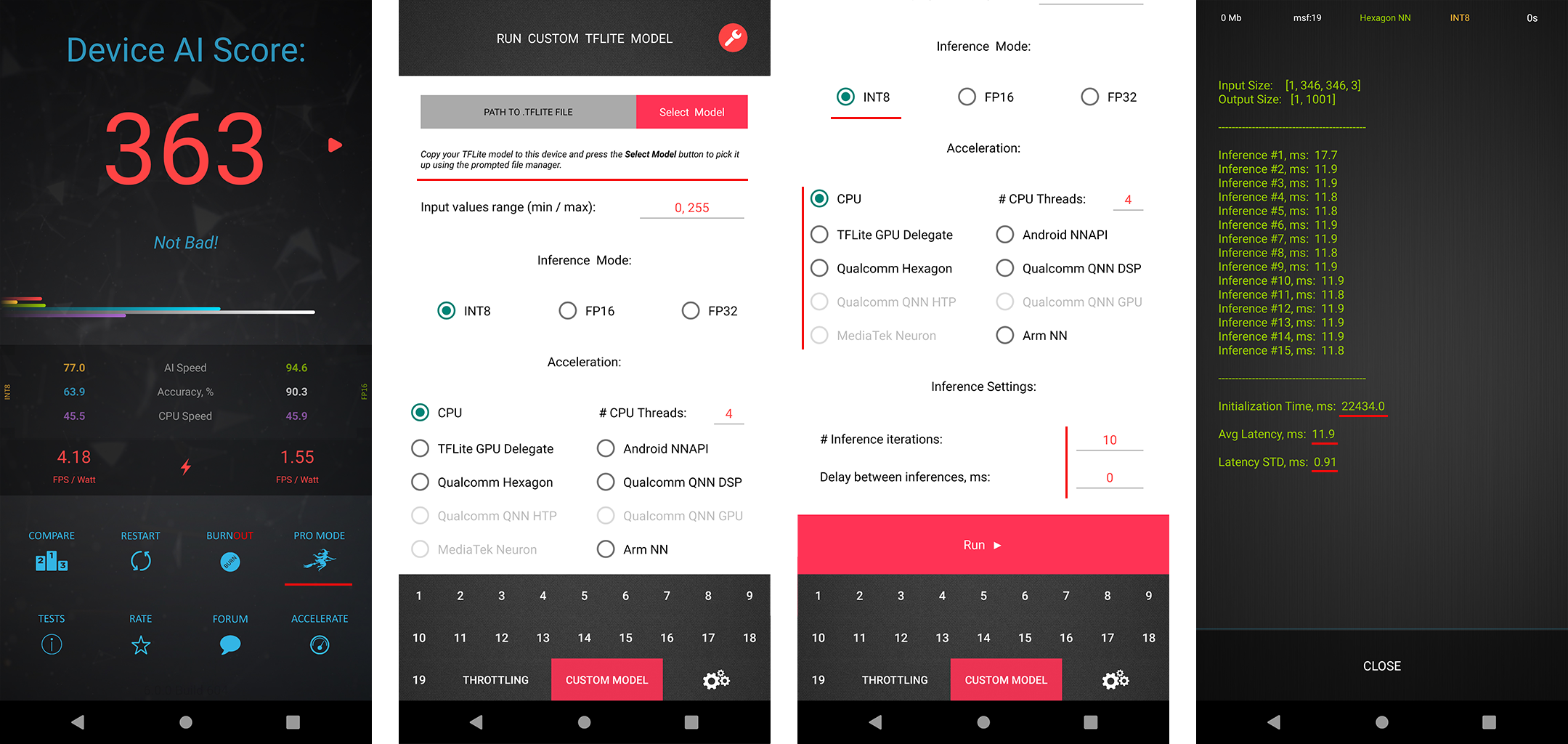}
		}
		\vspace{0.2cm}
		\caption{Loading and running custom TensorFlow Lite models with AI Benchmark application. The currently supported acceleration options include Qualcomm QNN, Hexagon NN, MediaTek Neuron, TFLite GPU, Android NNAPI and ARM NN delegates as well as CPU inference (LiteRT / XNNPACK backends). The latest app version can be downloaded at \url{https://ai-benchmark.com/download}}
		\label{fig:ai_benchmark_custom}
	\end{figure*}

	\section{Challenge}
	
	To design an efficient and practical solution for mobile-related tasks, the following key components are essential:
	
	
	\begin{enumerate}
		\item A high-quality and large-scale dataset that can be used to train and evaluate the solution;
		\item An efficient way to check the runtime and debug the model locally without any constraints;
	\end{enumerate}
	
	This challenge addresses the above issues. Real training data, tools, and runtime evaluation options provided to the challenge participants are described in the next sections.
	
	\subsection{Dataset}
	
In this challenge, we use a popular DPED dataset~\cite{ignatov2017dslr} that consists of photos taken in the wild synchronously by three smartphones and one DSLR camera. In total, the dataset provides over 22K photos, including 4549 photos from Sony smartphone, 5727 from iPhone and 6015 photos from each Canon and BlackBerry cameras. The dataset also provides aligned phone-DSLR paired patches of resolution 100$\times$100 px for training neural networks. In this challenge, we consider only a subtask of mapping photos from a very old iPhone 3GS device into the photos from Canon 70D DSLR, one such sample photo pair is shown in Fig.~\ref{fig:example_photos}.
	
	\subsection{Local Runtime Evaluation}
	
	When developing AI solutions for mobile devices, it is vital to be able to test the designed models and debug all emerging issues locally on available devices. For this, the participants were provided with the \textit{AI Benchmark} application~\cite{ignatov2018ai,ignatov2019ai} that allows users to load any custom TensorFlow Lite model and run it on any Android device with all supported acceleration options. This tool contains the latest versions of the \textit{Qualcomm QNN, MediaTek Neuron, Hexagon NN, TFLite GPU, Android NNAPI, Samsung ENN, Amlogic AML} and \textit{Arm NN} delegates, therefore supporting all current mobile platforms and providing the users with the ability to execute neural networks on smartphone NPUs, APUs, DSPs, GPUs and CPUs.
	
	\smallskip
	
	To load and run a custom TensorFlow Lite (LiteRT) model, one needs to follow the next steps:
	
	\begin{enumerate}
		\setlength\itemsep{0mm}
		\item Download AI Benchmark from the official website\footnote{\url{https://ai-benchmark.com/download}} or from Google Play\footnote{\url{https://play.google.com/store/apps/details?id=org.benchmark.demo}} and run its standard tests.
		\item After the end of the tests, enter the \textit{PRO Mode} and select the \textit{Custom Model} tab there.
		\item Put the exported TFLite (LiteRT) model into the \textit{Download} folder of the device.
		\item Select mode type \textit{(INT8, FP16, or FP32)}, the desired acceleration / inference options and run the model.
	\end{enumerate}
	
	\noindent These steps are also illustrated in Fig.~\ref{fig:ai_benchmark_custom}. The instructions on how to generate TFLite model from PyTorch, TensorFlow and Keras pre-trained models were provided in the MAI Workshop GitHub repository~\footnote{\url{https://github.com/aiff22/MAI-2025-Workshop}}.
	
	\subsection{Runtime Evaluation on the Target Platform}
	
	\begin{table*}[t!]
		\centering
		\resizebox{\linewidth}{!}
		{
			\begin{tabular}{l|c|cc|ccc|ccccc|c}
				\hline
				Team \, & \, Author \, & \, Framework \, & \, Model Size, \, & \, PSNR$\uparrow$ \, & \, SSIM$\uparrow$ \, & \, MOS$\uparrow$ \, & \multicolumn{2}{c}{\, Snapdragon 8 Elite Gen 5\,} & \multicolumn{2}{c}{\, Dimensity 9500\,} & Avg. GPU & \, Final Score \\
				& & & \small{KB} & & & & \, \small{GPU, ms $\downarrow$} \, & \, \small{NPU, ms $\downarrow$} \, &\, \small{GPU, ms $\downarrow$} \, & \, \small{NPU, ms $\downarrow$} \, &  \small{Runtime, ms $\downarrow$} & \\
				\hline
				\hline
				DaHua-IIG & morty0229 & PyTorch & 3240 & 22.20 & 0.7881 & 4.1 & 23.8 & 21.8 & 60.4 & \small{Failed} & 38 & 164 \\
				Capybara & genai4e\_cvpr2 & PyTorch & 3692 & 21.82 & 0.7653 & 3.2 & 291 & 87 & 266 & 73.7 & 278 & 3.8 \\
				Mivlab & yqh616 & PyTorch & 1127 & 22.19 & 0.7768 & 3.3 & 3373 & \small{Failed} & 5086 & \small{Failed} & 4142 & 0.49 \\
				Raytrontek Image & zjjforever & PyTorch & 3585 & 20.59 & 0.7380 & 2.3 & 113 & 39.2 & 146 & 29.6 & 128 & 0.43 \\
				DH-XHDL-Team & dzyxwyk & TensorFlow & 8662 & 20.55 & 0.7601 & 1.2 & 30.8 & 30.4 & 52.4 & 12.7 & 40 & 0.28 \\
				MotongAI & Alex\_Fall & PyTorch & 7344 & 22.17 & 0.7808 & 3.6 & 21119 & \small{Failed} & 22772 & \small{Failed} & 21930 & 0.14 \\
				\hline
			\end{tabular}
		}
		\vspace{2.6mm}
		\caption{\small{Mobile AI 2026 RGB Image Enhancement challenge results and final rankings. During the runtime measurements, the models were processing photos of resolution 1920$\times$1088 pixels on the Snapdragon 8 Elite Gen 5 and Dimensity 9500 SoCs. Team \textit{DaHua-IIG} is the challenge winner.}}
		\label{tab:results}
	\end{table*}
	
	In this challenge, we use two flagship mobile SoCs as our target evaluation platforms: \textit{Qualcomm Snapdragon 8 Elite Gen 5} and \textit{MediaTek Dimensity 9500} chipsets with the \textit{Adreno 840} and \textit{Arm Mali G1-Ultra MC12} GPUs, respectively, that can be found in numerous flagship smartphones. The models were parsed and accelerated using the TFLite GPU delegate~\cite{lee2019device} that demonstrated the best runtime results on both platforms. NPU runtimes provided for the reference were obtained by running the models with the Qualcomm QNN and MediaTek Neuron TFLite delegates on the Snapdragon 8 Elite Gen 5 and the Dimensity 9500 SoCs. All final solutions were tested using the aforementioned AI Benchmark application.
	
	\subsection{Challenge Phases}
	
	The challenge consisted of the following phases:
	
	\vspace{-0.8mm}
	\begin{enumerate}
		\item[I.] \textit{Development:} the participants get access to the data and AI Benchmark app, and are able to train the models and evaluate their runtime locally;
		\item[II.] \textit{Validation:} the participants can upload their models to the remote server to check the fidelity scores on the validation dataset, and to compare their results on the validation leaderboard;
		\item[III.] \textit{Testing:} the participants submit their final results, codes, TensorFlow Lite models, and factsheets.
	\end{enumerate}
	\vspace{-0.8mm}
	
	\subsection{Scoring System}
	
	All solutions were evaluated using the following metrics:
	
	\begin{itemize}
		\setlength\itemsep{-0.2mm}
		\item Peak Signal-to-Noise Ratio (PSNR) measuring fidelity score,
		\item Structural Similarity Index Measure (SSIM) as proxy for perceptual score,
		\item Mean Opinion Score (MOS) obtained from a user study and used to evaluate the perceived aesthetic quality of the enhanced images,
		\item The runtime on the target Snapdragon 8 Elite Gen 5 and Dimensity 9500 GPU.
	\end{itemize}
	
	The score of each final submission was evaluated based on the next formula ($C$ is a constant normalization factor):
	
	\smallskip
	\begin{equation*}
		\text{Final Score} \,=\, \frac{2^{2 \cdot \text{(PSNR + MOS)}}}{C \cdot \text{runtime}},
	\end{equation*}
	\smallskip
	
	During the final challenge phase, the participants did not have access to the test dataset. Instead, they had to submit their final TensorFlow Lite models that were subsequently used by the challenge organizers to check both the runtime and the fidelity results of each submission under identical conditions. This approach solved all the issues related to model overfitting, reproducibility of the results, and consistency of the obtained runtime / accuracy values.
	
	\section{Challenge Results}
	
	In this challenge, 6 teams entered the final phase and submitted their results, TFLite models, codes, executables and factsheets. Table~\ref{tab:results} summarizes the final challenge results for both tracks and report PSNR, SSIM and runtime numbers for the valid solutions on the final test dataset and on the target evaluation platform. The proposed methods are described in section~\ref{sec:solutions}, and the team members and affiliations are listed in Appendix~\ref{sec:apd:team}.
	\subsection{Results and Discussion}

	Almost all submitted solutions demonstrated very high efficiency: the majority of models are able to process one Full HD resolution (1920$\times$1088 px) image under 300 ms on the target mobile GPUs. Team \textit{DaHua-IIG} is the challenge winner: it achieved both the best perceptual image reconstruction results and an exceptionally good runtime of 23.8 ms and 60.4 ms on the Adreno and Arm GPUs, respectively. This team used a knowledge distillation strategy to overcome the misalignment of the input-target image pairs and trained a relatively shallow student CNN model for actual on-device inference. Several teams used ideas and blocks from transformer architectures in this challenge, however, despite their efficiency, they also significantly increased the latency of the resulting models.
	
	A notable progress that should be highlighted is the ease of model conversion to the target TFLite format. Table~\ref{tab:results} shows that the majority of solutions in this challenge were developed using the PyTorch ML framework. Due to the recently presented \textit{litert-torch} plugin~\footnote{\url{https://github.com/google-ai-edge/litert-torch}} developed by the LiteRT team, one can now directly convert the majority of PyTorch models to the TFLite format without the need for TensorFlow / Keras model re-implementation or use of ONNX as an intermediary format. This significantly simplifies the development process of mobile-friendly deep learning architectures for the majority of researchers using PyTorch as their main framework for AI model development and training.
	
	\section{Challenge Methods}
	\label{sec:solutions}
	
	\noindent This section describes solutions submitted by top teams participating in the final stage of the MAI 2026 RGB Image Enhancement challenge.
	
	\subsection{DaHua-IIG}
	
	\begin{figure*}[h!]
		\centering
		\setlength{\tabcolsep}{1pt}
		\resizebox{0.8\linewidth}{!}
		{
			\includegraphics[width=1.0\linewidth]{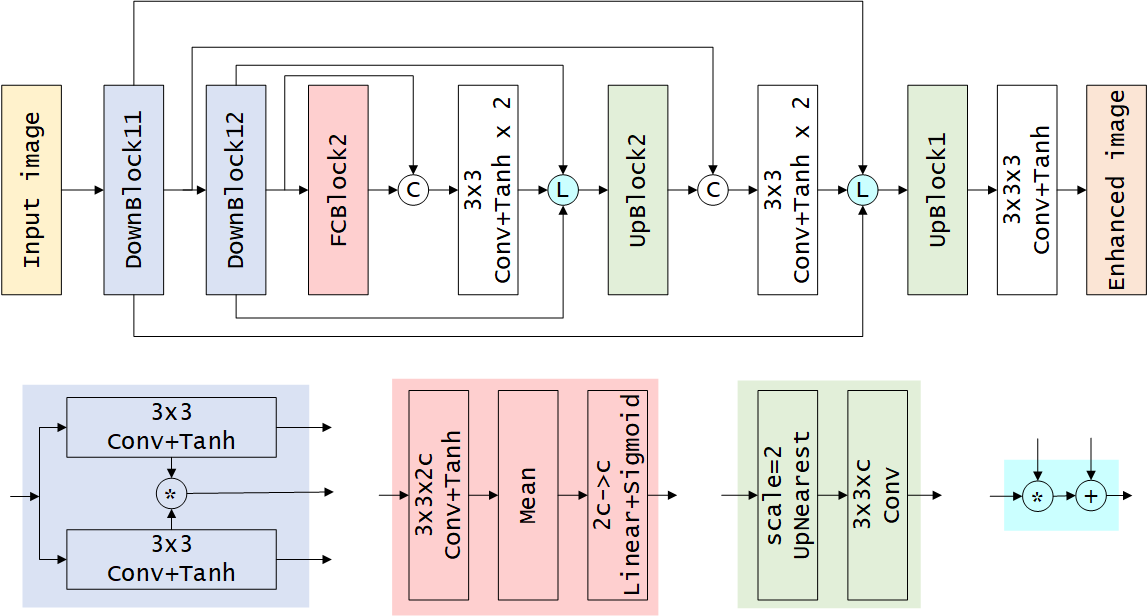}
		}
		\vspace{0.2cm}
		\caption{DENet architecture proposed by team DaHua-IIG}
		\label{fig:DaHua-IIG}
	\end{figure*}
	
	Team DaHua-IIG proposed the DENet architecture illustrated in Figure~\ref{fig:DaHua-IIG}. Since the input and ground-truth image pairs in the DPED dataset are not always perfectly aligned at the pixel level, the authors first trained a teacher model on this dataset with strong representational capacity, and then used the outputs of this model as pseudo ground truth for the student model~\cite{Hinton2015Distilling}. Through this distillation strategy they were able to achieve a favorable trade-off between the enhancement quality and computational efficiency of the student model.
	
	For the teacher model, the authors adopted the RetinexFormer~\cite{cai2023retinexformer} model that effectively integrates the Retinex theory with a transformer-based architecture. It achieves the state-of-the-art performance in low-light image enhancement, excelling in detail recovery and color fidelity, making it well-suited for the sRGB image enhancement task. As shown in Figure~\ref{fig:DaHua-IIG}, the student model is derived from a modified version of the PLUnet~\cite{ignatov2025MAIRGB}~--- the winning method in the 2025 sRGB image enhancement challenge, known for its outstanding performance in this domain. The authors added two specific modifications to the PLUnet. First, to further reduce the inference time, one \textit{FCBlock} was removed from the original architecture. Second, the authors applied channel pruning to decrease the model's computational complexity, making it more suitable for practical deployment. 
	
	During the training of both the teacher and student models, the authors employed a combination of Multi-Scale $L_2$ loss, Multi-Scale Structural Similarity (MS-SSIM) loss, and VGG-19-based Content loss. The Multi-Scale $L_2$ loss was defined as:
	\begin{equation*}
	\mathcal{L}_{MSMSE}\left( X,Y \right) =\sum_i{\mathcal{L}_{MSE}\left( X_i,Y_i \right)},
	\end{equation*}
	\begin{equation*}
	\mathcal{L}_{MSE}\left( X,Y \right) =\lVert X-Y \rVert _{2}^{2},
	\end{equation*}
	where $X_i$ and $Y_i$ are the representations of the image pair at different bilinear sampling scales $i \in$ [1, 0.5, 0.25].
	For the teacher model, the weighting factors for these three losses were set to 1, 500, and 1, respectively. For the student model, the corresponding weights were 1, 400, and 10.
	
	The parameters of the teacher model were optimized using the Adam optimizer ($\beta_{1}$ = 0.9 and $\beta_{2}$ = 0.999) with a batch size of 32. The learning rate was initially set to 2$\times$$10^{-4}$ and then gradually reduced to 1$\times10^{-6}$ by the cosine annealing scheme after 1000000 iterations. For training the student model, the batch size was set to 96, and the learning rate was initially set to  1$\times10^{-3}$.
	
	\subsection{Capybara}
	
	\begin{figure}[h!]
		\centering
		\setlength{\tabcolsep}{1pt}
		\resizebox{1.0\linewidth}{!}
		{
			\includegraphics[width=1.0\linewidth]{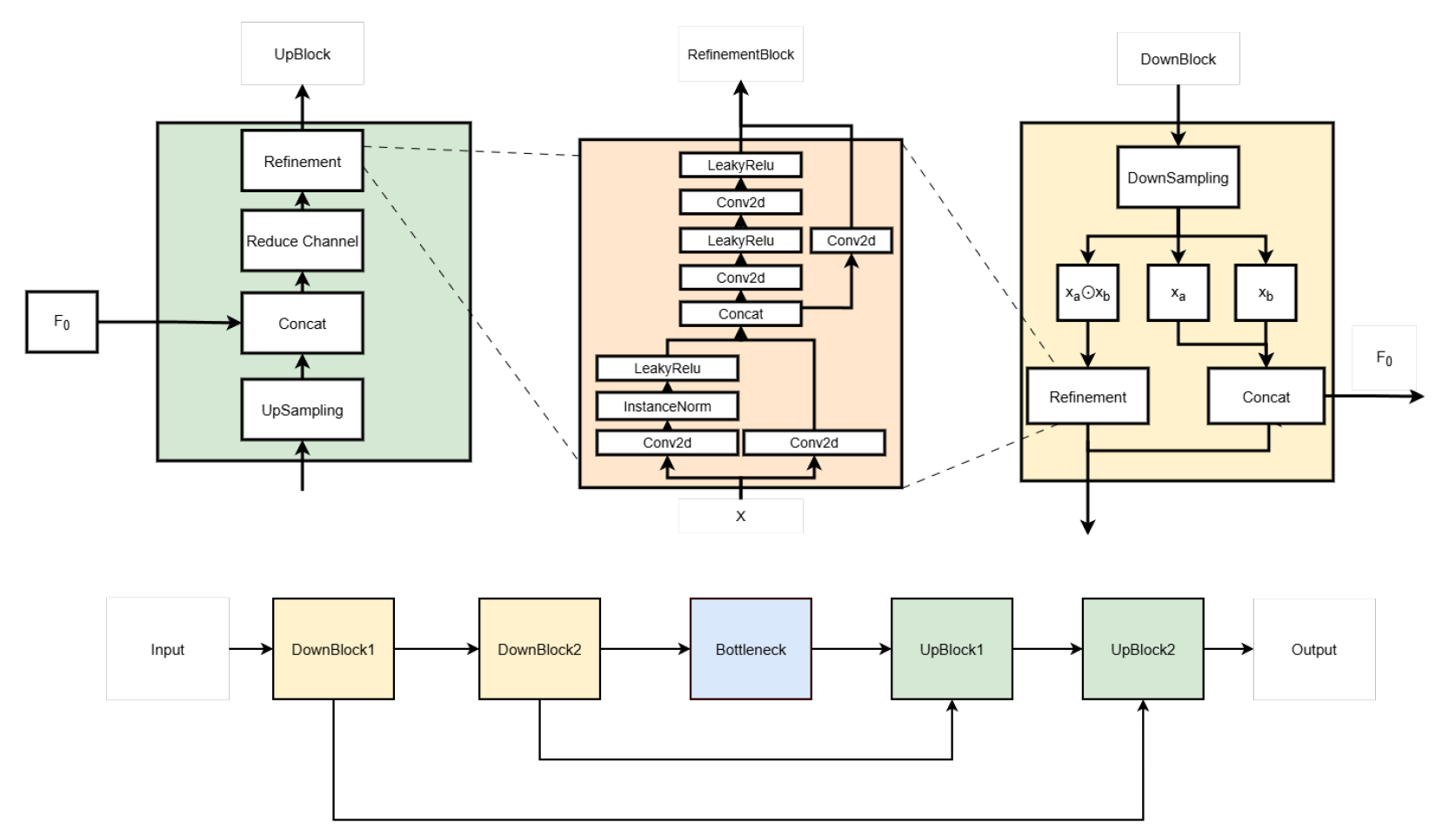}
		}
		\vspace{0.2cm}
		\caption{Model architecture proposed by team Capybara}
		\label{fig:Capybara}
	\end{figure}
	
	Team Capybara proposed a hybrid architecture consisting of a lightweight gated multi-scale encoder-decoder and a set of UNet-style refinement blocks (Fig.~\ref{fig:Capybara}). The encoder-decoder branch provides efficient feature extraction, whereas the refinement blocks improve detail reconstruction, color restoration, and structural consistency at multiple spatial scales~\cite{tinh2026bridging}.
	
	The training objective is defined as a weighted combination of three loss functions: 1) a PSNR-based loss that is computed on the luminance channel Y in order to emphasize perceptual brightness fidelity; 2) MS-SSIM loss, and 3) \textit{OutlierAwareLoss} used to reweight pixel-wise errors according to per-image deviation statistics, thereby assigning greater emphasis to more challenging regions.
	
	The model was implemented in PyTorch Lightning, its parameters were optimized using Adam with a learning rate of 10e--4. For scheduling, cosine annealing with warm restarts was adopted.
		
	\subsection{Mivlab}
	
	\begin{figure}[h!]
		\centering
		\setlength{\tabcolsep}{1pt}
		\resizebox{1.0\linewidth}{!}
		{
			\includegraphics[width=1.0\linewidth]{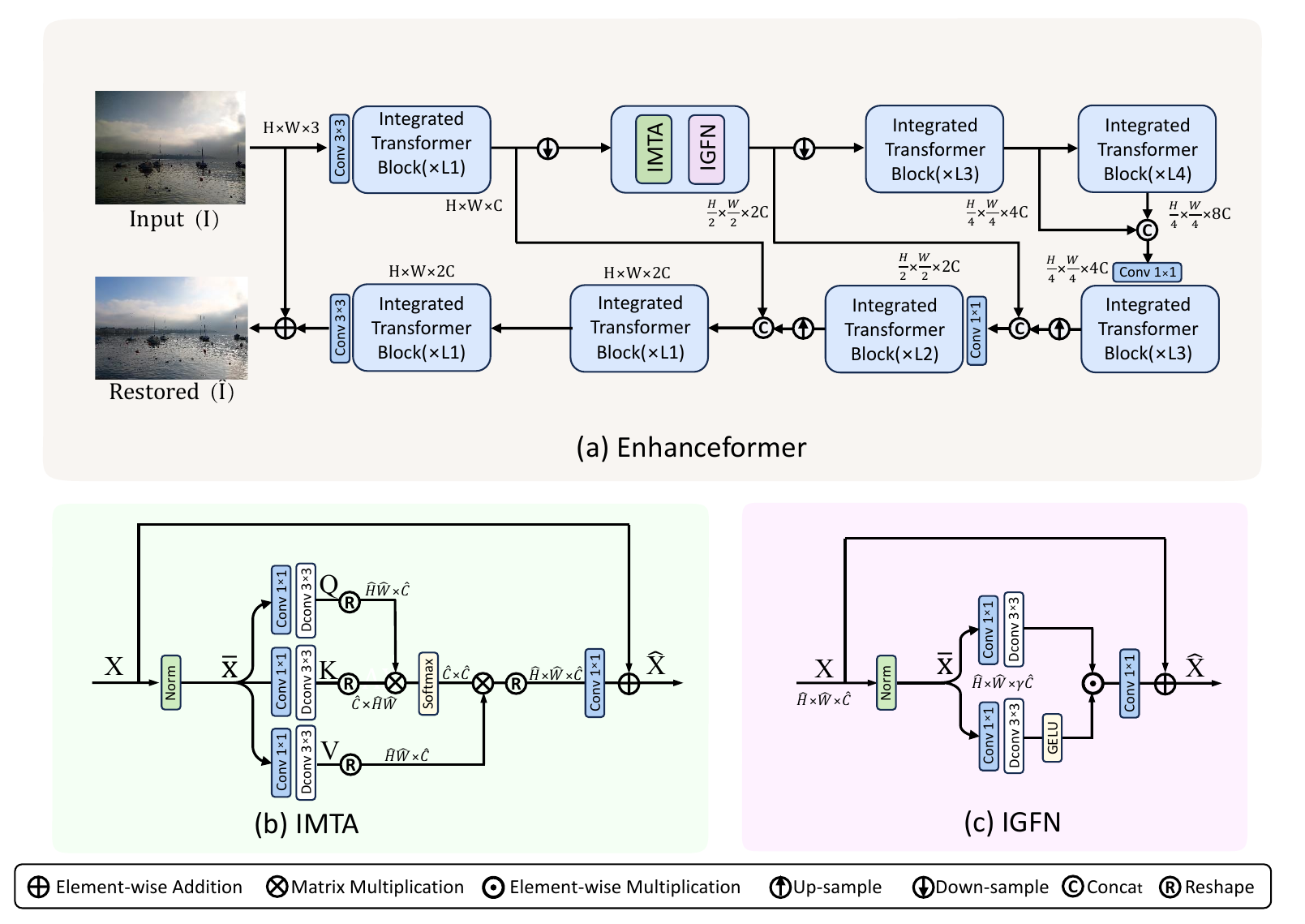}
		}
		\vspace{0.2cm}
		\caption{Enhanceformer model proposed by team Mivlab}
		\label{fig:Mivlab}
	\end{figure}
	
	Team Mivlab presented the Enhanceformer model which architecture is demonstrated in Figure~\ref{fig:Mivlab}. It adopts a 4-level encoder-decoder U-Net like architecture that uses transformer blocks and transposed attention modules to reduce the number of parameters and the model's computational complexity. From level 1 to level 4, the number of \textit{Integrated Transformer Blocks} are [1, 1, 1, 2], the number of attention heads in \textit{IMTA} are [1, 2, 4, 8], and the number of channels are [8, 16, 32, 64]. The refinement stage contains 1 blocks. The channel expansion factor in \textit{GDFN} is $\gamma = 2.66$.
	
	The model was trained to minimize $L_1$ loss with the AdamW optimizer ($\beta_1$ = 0.9, $\beta_2$ = 0.999, weight decay 1e--4) for 510K iterations with the initial learning rate 3e--4 gradually reduced to 1e--6 with the cosine annealing strategy. The model was trained on patches of size 128$\times$128 with a batch size of 960. For data augmentation, the authors used horizontal and vertical flips.
	
	\subsection{MotongAI}
	
	\begin{figure}[h!]
		\centering
		\setlength{\tabcolsep}{1pt}
		\resizebox{1.0\linewidth}{!}
		{
			\includegraphics[width=1.0\linewidth]{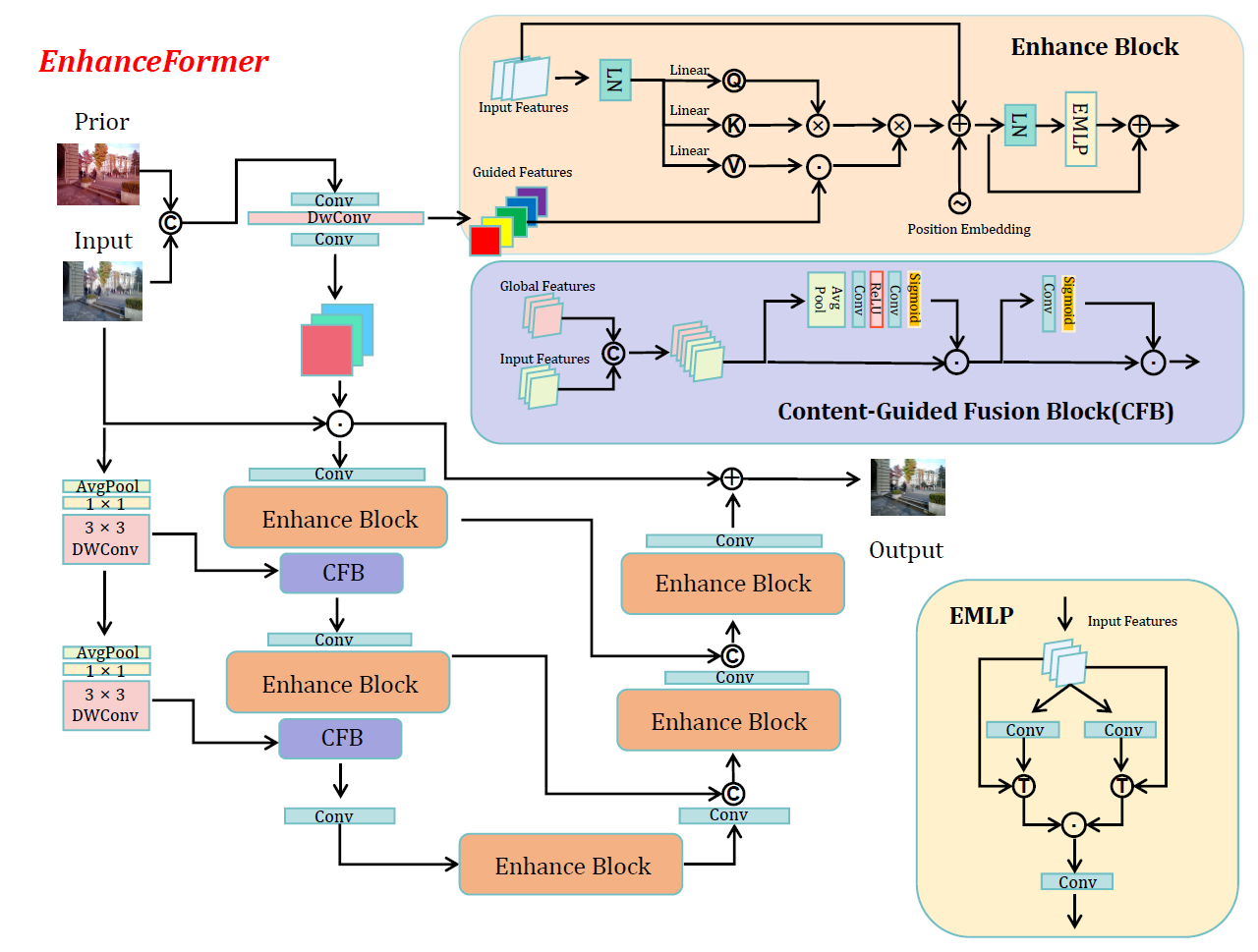}
		}
		\vspace{0.2cm}
		\caption{Model architecture proposed by team MotongAI}
		\label{fig:MotongAI}
	\end{figure}
	
	Team MotongAI presented the Enhanceformer model which architecture is demonstrated in Figure~\ref{fig:MotongAI}. Its design is also inspired by the Retinexformer~\cite{cai2023retinexformer} architecture. The model has a U-Net-like structure and consists of multiple \textit{Enhance Blocks} and \textit{Content-Guided Fusion Block} (CFB) processing the image at different scales. Each \textit{Enhance Block} consists of a \textit{LayerNorm}, a self-attention mechanism and an \textit{EMLP} block. The \textit{CFB} block effectively combines the extracted global information with the input features at both channel attention and spatial attention scales.
		
	The model was trained to minimize $L_1$ loss using the Adam optimizer with $\beta_1$ = 0.9, $\beta_2$ = 0.99, and a batch size of 512. Cosine annealing learning rate scheduler was employed to control the variation of the learning rate. The initial learning rate was set to 1e--3, decreasing during the training to 5e--6.

	\section{Additional Literature}
	
	An overview of the past challenges on mobile-related tasks together with the proposed solutions can be found in the following papers:
	
	\begin{itemize}
		\item Image Super-Resolution:\, \cite{ignatov2018pirm,cai2019ntire,timofte2018ntire,lugmayr2020ntire}
		\item Learned End-to-End ISP:\, \cite{ignatov2019aim,ignatov2020aim,wirzberger2022lan,ignatov2022microisp,ignatov2022pynetv2}
		\item RGB Image Enhancement:\, \cite{ignatov2018pirm,ignatov2019ntire,ignatov2017dslr,ignatov2018wespe}
		\item Bokeh Effect Rendering:\, \cite{ignatov2019aimBokeh,ignatov2020aimBokeh,ignatov2020rendering}
		\item Image Denoising:\, \cite{abdelhamed2019ntire,abdelhamed2020ntire,flepp2024real}
	\end{itemize}
	
	\section*{Acknowledgements}
	
	We thank AI Witchlabs, ETH Z\"urich, and the University of W\"urzburg (Computer Vision Lab), the organizers and sponsors of this Mobile AI 2026 challenge.
	
	\appendix
	\section{Teams and Affiliations}
	\label{sec:apd:team}
	
	\bigskip
	
	\subsection*{Mobile AI 2026 Team}
	\noindent\textit{\textbf{Title: }}\\ Mobile AI 2026 RGB Photo Enhancement Challenge \\
	\noindent\textit{\textbf{Members:}}\\ Andrey Ignatov$^{1,3}$ \textit{(andrey@vision.ee.ethz.ch)}, Georgy Perevozchikov$^2$ \textit{(georgii.perevozchikov@uni-wuerzburg. de)}, Radu Timofte$^{2,3}$ \textit{(radu.timofte@uni-wuerzburg.de)}\\
	\noindent\textit{\textbf{Affiliations: }}\\
	$^1$ Computer Vision Lab, ETH Zurich, Switzerland\\
	$^2$ Computer Vision Lab, University of W\"urzburg, Germany\\
	$^3$ AI Witchlabs, Switzerland\\
	
	\subsection*{DaHua-IIG}
	\noindent\textit{\textbf{Title:}}\\ Distillation-Based Efficient Network for sRGB Image Enhancement on Smartphones \\
	\noindent\textit{\textbf{Members: }}\\ Dong Zhang \textit{(zxdwe@qq.com)}, Zhao Ran, Yang Zhang, Xiaochen Li, Shichang Ju, Yiqin Cao, Yang Lu\\
	\noindent\textit{\textbf{Affiliations: }}\\
	Zhejiang Dahua Technology Co., Ltd.\\
	
	\subsection*{Capybara}
	\noindent\textit{\textbf{Title:}}\\ Bridging the Training–Deployment Gap: Gated Encoding and Multi-Scale Refinement for Efficient Quantization-Aware Image Enhancement~\cite{tinh2026bridging}\\
	\noindent\textit{\textbf{Members: }}\\ Nguyen Nhu Tinh Anh$^1$ \textit{(anh.nguyennhu2306@hcmut. edu.vn)}, To Thanh Dat$^1$, Nguyen Ngoc Ky Duyen$^2$, Nguyen Trong Nghia$^1$, Bui Minh Hieu$^3$\\
	\noindent\textit{\textbf{Affiliations: }}\\
	$^1$ Ho Chi Minh City University of Technology - VNUHCM\\
	$^2$ Carleton University\\
	$^3$ Da Nang University of Economics (DUE)\\
	
	\subsection*{Mivlab}
	\noindent\textit{\textbf{Title:}}\\ Lightweight image enhancement network based on transformer \\
	\noindent\textit{\textbf{Members: }}\\ Qiuhai Yan \textit{(yanqiuhai16@163.com)}, Sunzhe Yang, Kang Chen, Tingting Wang, Faming Fang\\
	\noindent\textit{\textbf{Affiliations: }}\\
	East China Normal University\\
	
	\subsection*{Raytrontek Image}
	\noindent\textit{\textbf{Title:}}\\ A Lightweight Mobile Image Enhancement Network \\
	\noindent\textit{\textbf{Members: }}\\ Jinjie Zhou \textit{(1943035411@qq.com)}, Yanglu Fengduo\\
	\noindent\textit{\textbf{Affiliations: }}\\
	Raytron Technology\\
	
	\subsection*{DH-XHDL-Team}
	\noindent\textit{\textbf{Members: }}\\ MoHao Wu \textit{(email:1025331980@qq.com)}, Lin Wang\\
	\noindent\textit{\textbf{Affiliations: }}\\
	Zhejiang Dahua Technology Co., Ltd.\\
	
	\subsection*{MotongAI}
	\noindent\textit{\textbf{Members: }}\\ Runhua Deng \textit{(2312666252@qq.com)}, Zhifeng Wang, Long Peng, Aiwen Jiang \\
	\noindent\textit{\textbf{Affiliations: }}\\
	Independent Researchers\\

	{\small
		\bibliographystyle{ieeenat_fullname}
		\bibliography{egbib}
	}
	
\end{document}